\documentclass[journal,twoside,web]{ieeecolor}
\usepackage{tmi}
\usepackage{cite}
\usepackage{amsmath,amssymb,amsfonts}
\usepackage{algorithmic}
\usepackage{graphicx}
\usepackage{textcomp}
\usepackage{multirow}
 \usepackage{color}
\usepackage{array}
\usepackage{utfsym}
\def\BibTeX{{\rm B\kern-.05em{\sc i\kern-.025em b}\kern-.08em
    T\kern-.1667em\lower.7ex\hbox{E}\kern-.125emX}}
\begin{document}
\title{Video-Instrument Synergistic Network for Referring Video Instrument Segmentation in Robotic Surgery}
\author{Hongqiu Wang, \IEEEmembership{Student Member, IEEE}, Lei Zhu, Guang Yang, \IEEEmembership{Senior Member, IEEE}, Yike Guo,\\ Shichen Zhang, Bo Xu, Yueming Jin, \IEEEmembership{Member, IEEE}
\vspace{-10mm}
\thanks{Hongqiu Wang, Lei Zhu, Shichen Zhang: Department of Systems Hub, Hong Kong University of Science and Technology (Guangzhou), China (e-mail: hwang007@connect.hkust-gz.edu.cn, leizhu@ust.hk, szhang213@connect.hkust-gz.edu.cn).}
\thanks{Guang Yang: Bioengineering/Imperial-X, Imperial College London, UK (e-mail: g.yang@imperial.ac.uk).}
\thanks{Yike Guo: Imperial College London and Hong Kong University of Science and Technology, UK and China (e-mail: yikeguo@ust.hk).}
\thanks{Bo Xu: Department of Anesthesiology, The General Hospital of Southern Theatre Command of PLA, Guangzhou, China (e-mail: xubo333@hotmail.com).}
\thanks{Yueming Jin: Department of Biomedical Engineering, National University of Singapore, Singapore (e-mail: ymjin@nus.edu.sg).}}

\maketitle
\begin{abstract}
Robot-assisted surgery has made significant progress, with instrument segmentation being a critical factor in surgical intervention quality. It serves as the building block to facilitate surgical robot navigation and surgical education for the next generation of operating intelligence.
Although existing methods have achieved accurate instrument segmentation results, they simultaneously generate segmentation masks for all instruments, without the capability to specify a target object and allow an interactive experience.
This work explores a new task of Referring Surgical Video Instrument Segmentation (RSVIS), which aims to automatically identify and segment the corresponding surgical instruments based on the given language expression.
To achieve this, we devise a novel Video-Instrument Synergistic Network (VIS-Net) to learn both video-level and instrument-level knowledge to boost performance, while previous work only used video-level information.
Meanwhile, we design a Graph-based Relation-aware Module (GRM) to model the correlation between multi-modal information (i.e., textual description and video frame) to facilitate the extraction of instrument-level information.
We are also the first to produce two RSVIS datasets to promote related research. Our method is verified on these datasets, and experimental results exhibit that the VIS-Net can significantly outperform existing state-of-the-art referring segmentation methods. Our code and our datasets will be released upon the publication of this work.
\vspace{-3mm}
\end{abstract}

\section{Introduction}
\label{sec:introduction}
\begin{figure}[t]
    \centering
    \includegraphics[width=0.5\textwidth]{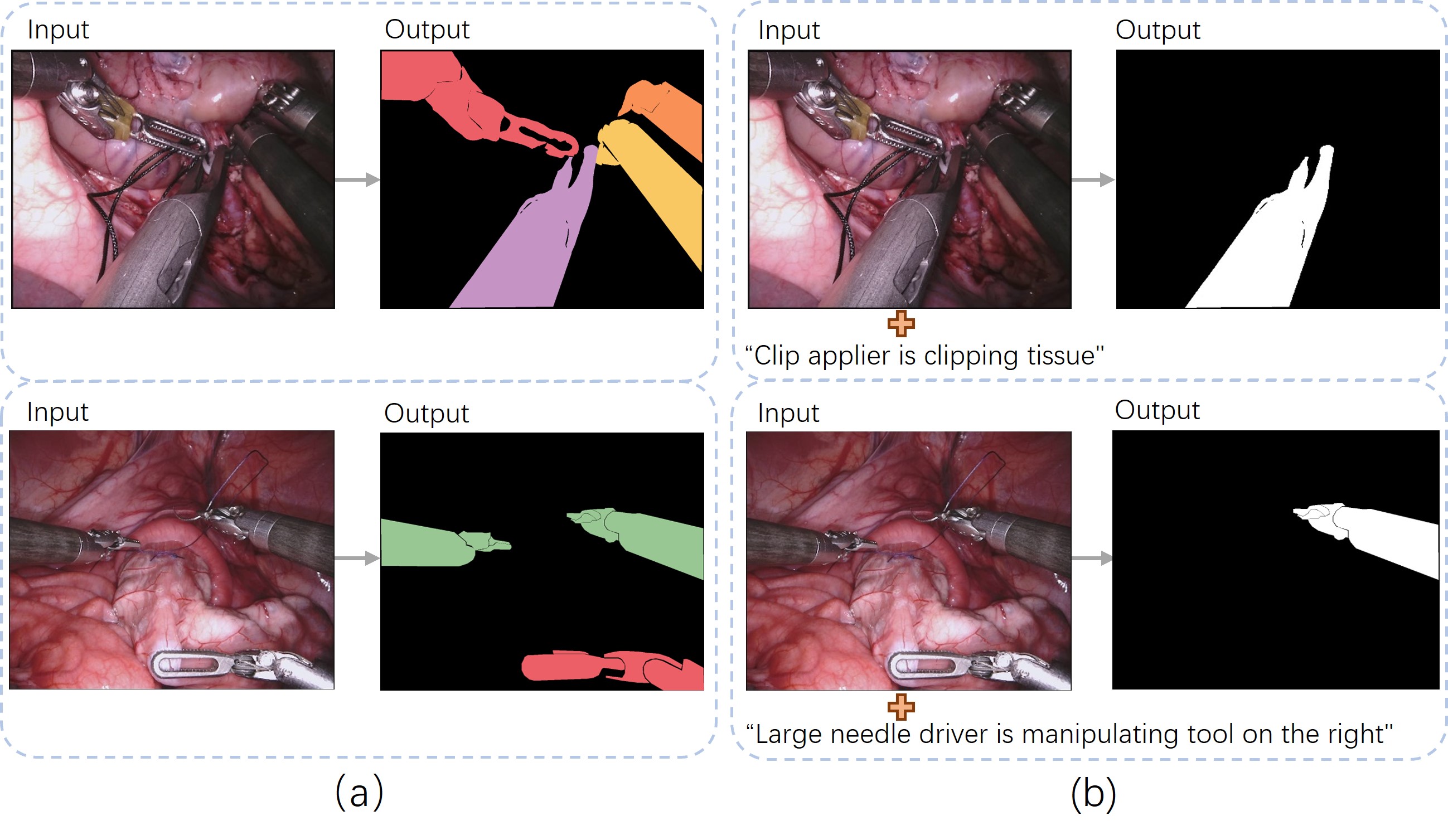}
    \vspace{-8mm}
    \caption{Comparison of (a) existing instrument segmentation task and (b) our referring surgical video instrument segmentation. The first column shows that we can select the main operating instrument of interest from complex scenes. The second column demonstrates that RSVIS can segment the instrument of interest based on the description when multiple instruments of the same type existing in one surgical scene.}
    \label{fig:f1}
    \vspace{-5mm}
\end{figure}

\IEEEPARstart{S}{urgical} instrument segmentation, which can identify different types of instruments used in the surgical procedure, has attracted wide attention due to its essentially foundational role for many downstream applications\cite{ross2021comparative,bouget2017vision,yang2022drr}, such as tool pose estimation \cite{sarikaya2017detection}, trajectory prediction \cite{osa2017online}, and even facilitating robotic navigation \cite{zang2019extremely} and task automation \cite{allan20192017,allan20202018,wang2023dynamic}. Additionally, it can promote cognitive assistance to surgeon perception \cite{gao2021future}. Embedding it in the augmented reality environment can bring new possibilities for the next generation of surgery education \cite{cao2019virtual}. Typically, existing methods focus on achieving real-time inference, by model light-weighting \cite{ni2020attention,shen2023branch}, and utilizing fewer annotations while maintaining accurate segmentation by semi-supervised or unsupervised learning \cite{sestini2023fun,garcia2021image,colleoni2022ssis}.
However, these methods only generate the segmentation mask of all instruments (cf. Fig. \ref{fig:f1} (a)) without the capability of specifying the one of interest, e.g., the primary operating instrument in the current surgical phase. 

A more desired and flexible situation is that given the reference information (e.g., text description), the system can automatically identify and segment the corresponding instruments (cf. Fig. \ref{fig:f1} (b)). Utilizing text for referring segmentation inherently provides a more natural way of human-computer interaction. Together with an AR system, it has the potential to enhance surgical education significantly \cite{kovoor2021validity,sheik2019next}. Surgical trainees can specify areas of interest and allow the AR system to overlay the referred interests onto the surgical scene, providing them with a more immersive, interactive, and personalized learning experience. Meanwhile, such a higher-level scene understanding shows the great potential for benefiting intra-operation, where the system can prioritize areas of interest as indicated by the surgeons, reduce computational complexity, and facilitate accurate, real-time and personalized robotic surgery navigation. 

With the efficacy to facilitate interaction and great potential to benefit a wide range of downstream tasks, the referring segmentation has emerged in the computer vision community \cite{liu2021cross,seo2020urvos}. The most related task is named referring video object segmentation \cite{gavrilyuk2018actor}, which aims to segment the object referred by natural language expressions in a natural video frame. The existing approaches \cite{liu2021cross,seo2020urvos} generally incorporate the vision and language features by employing attention mechanisms to align video with text and then adopt a decoder to generate corresponding object masks. 
However, considering the intricate and ever-evolving nature of surgical environments, compounded by the significant resemblance often shared among various surgical instruments, these methods are insufficient to tackle the referring segmentation task in robotic surgery scenarios.

We identify two main limitations of existing methods hindering them from accurately referring surgical instrument segmentation. First, these methods only extract overall features from the original video (video-level) and then combine text to generate predictions. Since the model requires performing pixel-wise segmentation of the referred object, the additional instrument-level representations of the referred object can contribute to the segmentation.
Second, referring segmentation requires a comprehensive understanding of the cross-modal sources (i.e., vision and language) to effectively focus on the objects to be segmented. In a surgical scenario with a multiplicity of instruments and where the instruments in particular have a high degree of similarity, it is crucial to accurately model the non-Euclidean distance between linguistic features and visual features of multiple instruments. 

In this paper, we take the \emph{first} step to embark on an essential video-language task in the surgical domain, named \emph{Referring Surgical Video Instrument Segmentation} (RSVIS), which can segment the text-referred instrument in the frame given a surgical procedure. To tackle this task, we construct two new comprehensive RSVIS datasets by re-annotating existing surgical instrument segmentation datasets from the well-known public benchmarks, MICCAI EndoVis17 and EndoVis18 Challenges \cite{allan20192017,allan20202018}. In alignment with the guidance provided by experienced surgeons, we designed a diverse array of video-text pairs for the numerous instruments employed in the complex and challenging surgical scenario. 
Drawing on the distinctive characteristics of the task, we propose a novel \emph{Video-Instrument Synergistic Network} (VIS-Net) to leverage both video-level and instrument-level information to improve the accuracy of referred segmentation. 
Furthermore, given that RSVIS involves the integrated understanding and the interaction of information from different modalities and drawing insights from the impressive ability of graph neural networks (GNNs) to capture non-Euclidean relationships between entities \cite{cui2022braingb,wang2018zero,zhu2023data}, we formulate a novel multi-modal GNN named the Graph-based Relation-aware Module (GRM) to capture the correlation between different modalities (i.e., text feature and image features of each instrument), to facilitate the extraction of instrument-level information. 
Experimental results demonstrate that our VIS-Net clearly outperforms state-of-the-art referring segmentation models. We shall release our code and dataset to facilitate future studies on the significant topic. Main contributions are summarized as follows: 

\begin{itemize}
\item To the best of our knowledge, we take the first step to explore the crucial interactive segmentation task in surgical scenarios, which can contribute to the development and integration of future context-aware intelligent systems in the operating room.
\item We introduce a Video-Instrument Synergistic Network, designed to enhance performance by effectively capturing and leveraging both video-level and instrument-level representations.
\item We devise a Graph-based Relation-aware Module to effectively capture the non-Euclidean distances in linguistic and distinct visual features of various instruments, bolstering the extraction of text-related visual embedding.
\item We construct two RSVIS datasets following the advice from our collaborative surgical team. Both will be released to facilitate and advance research in the field of surgical video language learning. We benchmark our method and other state-of-the-art approaches in the datasets. Experimental results demonstrate the superiority of our method to others.
\end{itemize}

\section{Related work}
\subsection{Semantic Segmentation of Surgical Scene}
Various advancements have been made in semantic segmentation of surgical surgery, incorporating various techniques and methodologies, mainly including combining segmentation networks with attention mechanisms \cite{ni2020attention,shen2023branch} and exploring some synthetic data for semi-supervised training \cite{sestini2023fun,garcia2021image,wang2022rethinking}. Ni \textit{et al.} propose an attention-guided lightweight network, utilizing depth-wise separable convolution as the basic unit to reduce computational costs, thereby performing surgical instruments segmentation in real-time \cite{ni2020attention}. 
Shen \textit{et al.} employ a lightweight encoder and branch aggregation attention mechanism to remove noise caused by reflection, water mist to improve segmentation accuracy and achieve a lightweight model \cite{shen2023branch}. Sestini \textit{et al.} design a fully unsupervised method for segmentation of binary surgical instruments relying only on implicit motion information and a priori knowledge of the shape of the instrument \cite{sestini2023fun}. Luis \textit{et al.} facilitate this task by generating large amounts of trainable data by synthesizing surgical instruments with real surgical backgrounds \cite{garcia2021image}. Wang \textit{et al.} utilize a single surgical background image and a few open-source instrument images to synthesize amounts of image variations for training \cite{wang2022rethinking}. However, previous approaches mainly focus on real-time or semi-supervised, unsupervised learning, and there has been no research on language-based interactive segmentation. Moreover, their methods cannot distinguish multiple instruments of the same type in a scene, so we explored a new interactive segmentation task in surgery, referring instrument segmentation.

\subsection{Referring Segmentation}
The objective of referring segmentation is to accurately delineate the specific object mentioned in a natural language expression within a still image, a composite task incorporating computer vision and natural language processing. This is initially introduced by Hu \textit{et al.} \cite{hu2016segmentation}, and they employ the FCN \cite{long2015fully} to extract visual features and LSTM \cite{hochreiter1997long} to extract language features. These features are fused to predict the segmentation mask of the referred object. Referring segmentation is then extended from single images to the domain of videos, where the segmentation of the correct target in dynamic video requires both motion and appearance information \cite{carreira2017quo,gavrilyuk2018actor}. Ding \textit{et al.} \cite{ding2022language} utilize the computed frame difference to extract the temporal information and the frame images to extract the visual information, and then take the language as the bridge to interact with the two types of information to accomplish this task. Liu \textit{et al.} \cite{liu2021cross} propose a progressive understanding scheme across modalities to align multi-modal features in multiple stages based on the expression. Besides, a multi-modal Transformer model treats the task as a sequence prediction problem by generating prediction sequences for all video objects and identifying the referred object \cite{botach2022end}. This design enables data association across multiple frames (tracking) to handle occlusions and motion blur. Nevertheless, previous approaches almost extract visual information on the video-level. Given that the segmentation task is a dense prediction task, we devise an instrument-level appearance feature embedding extraction branch to boost the accuracy of boundary segmentation.

\begin{figure*}[h!]
    \centering
    \includegraphics[width=0.96\textwidth]{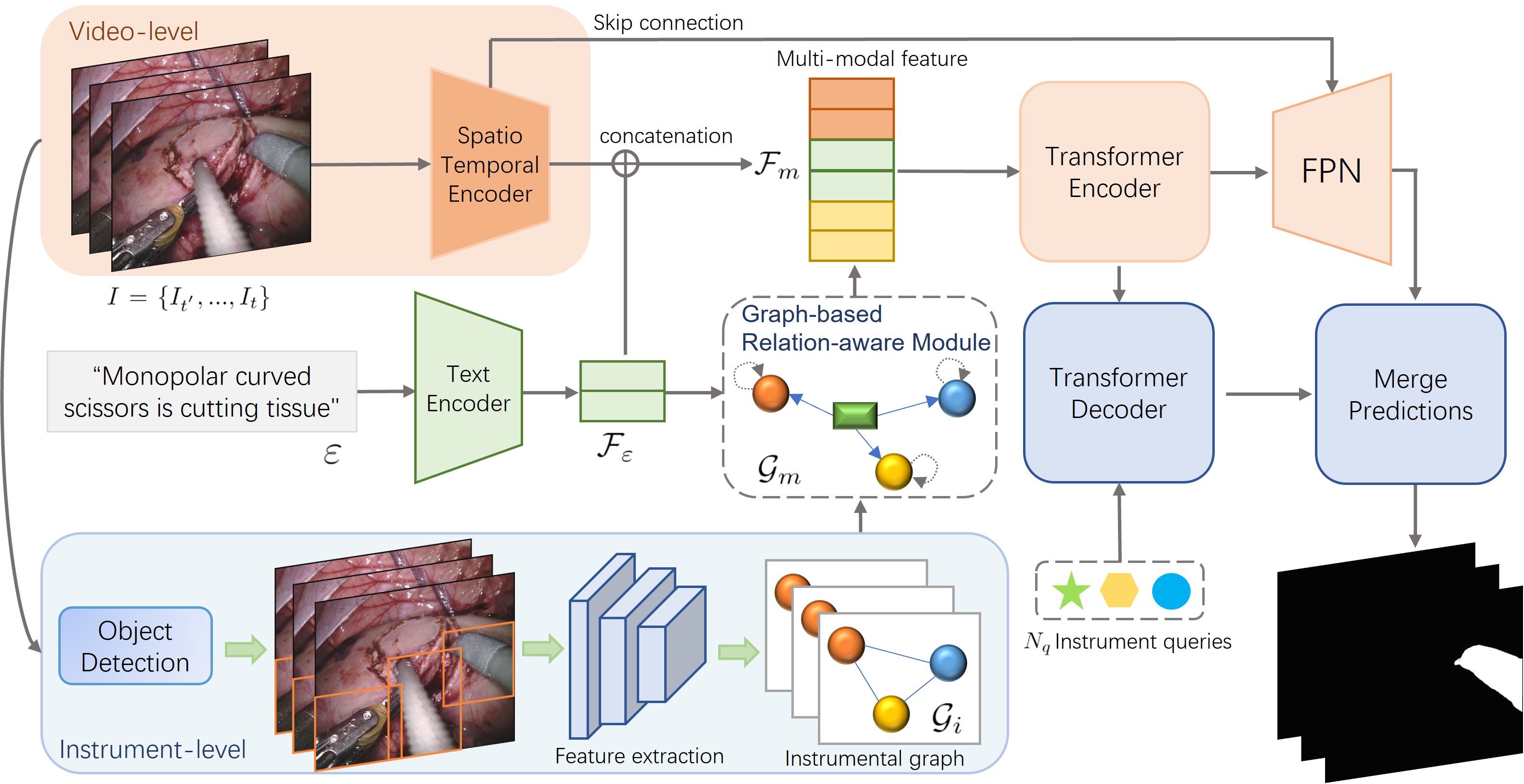}
    \caption{Overview of VIS-Net. Given a video clip $I$ accompanied by its text description $\varepsilon$, our VIS-Net extracts the visual embeddings in both video-level and instrument-level, and the textual embedding. We then devise GRM with a multi-modality directed graph $\mathcal{G}_m$, to enhance instrument-level embeddings by aggregating the knowledge guided by the text. We integrate the augmented instrument-level embeddings with the other two, and the obtained $\mathcal{F}_m$ is fed into a Transformer encoder, followed by an FPN and a decoder to predict the final segmentation result.}
    \label{fig:Overview}
    \vspace{-3mm}
\end{figure*}

\begin{figure}[h!]
    \centering
    \includegraphics[width=0.48\textwidth]{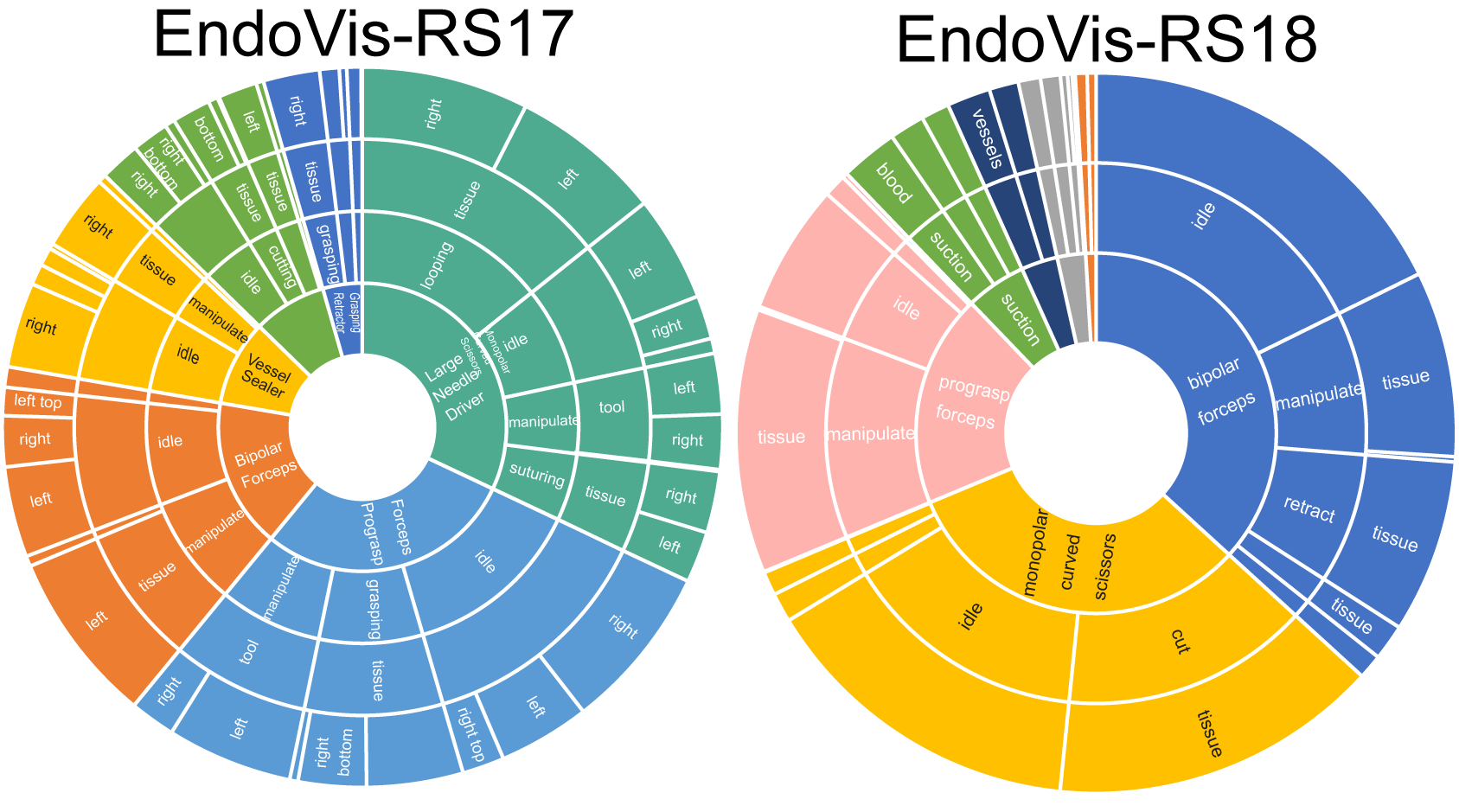}
    \vspace{-2mm}
    \caption{Distribution of textual descriptions in the two datasets categorized by their primary words. Words are ordered radially, emanating from the center.}
    \label{fig:data}
    \vspace{-3mm}
\end{figure}

\subsection{Segmentation Datasets in Robotic Surgery}
There are several publicly available robotic surgery datasets that have significantly contributed to relevant segmentation research \cite{allan20192017,allan20202018,garcia2021image,maier2021heidelberg,ross2021comparative}. EndoVis 2017 Challenge dataset consists of $8\times225$ and $8\times75$ frames of stereo camera image video acquired from the da Vinci Xi robot during several different porcine surgeries \cite{allan20192017}. The annotations of the dataset include binary instrument segmentation, instrument part segmentation, and instrument type segmentation. EndoVis 2018 Challenge dataset comprises 16x149 robotic nephrectomies and provided full scene segmentation with labeling \cite{allan20202018}. Pfeiffer \textit{et al.} \cite{pfeiffer2019generating} employ unpaired image-to-image translation techniques to generate a large-scale annotated dataset for liver segmentation in laparoscopic images, using the Cholec80 dataset \cite{twinanda2016endonet} as the source of the real-world data. González \textit{et al.} \cite{gonzalez2020isinet} manually expand the annotations of the EndoVis 2018 dataset and introduce additional instrument instance information for the instance segmentation in surgical scenes. Roß \textit{et al.} \cite{ross2021comparative} and Maier-Hein \textit{et al.} \cite{maier2021heidelberg} provide the ROBUST-MIS dataset comprising 10,000 challenging images from three various surgery types to validate the robustness and generalization capabilities of the binary and multiple instance segmentation methods. The CholecSeg8k dataset \cite{hong2020cholecseg8k} comprises endoscopic images annotated with semantic segmentation labels, and is constructed based on the Cholec80 dataset \cite{twinanda2016endonet}. The CaDIS dataset \cite{grammatikopoulou2021cadis} is a comprehensive collection of 4,670 surgical microscope images, each of which is annotated with pixel-wise labels for anatomical structures and surgical instruments used in cataract surgery procedures. 
However, previous datasets do not provide textual descriptions paired with segmented instruments, which also means that multiple instruments with the same type in one surgical scene cannot be distinguished. Therefore, we take the lead in annotating the paired textual description based on the well-known Challenge datasets of EndoVis 2017 and 2018 with the help of surgeons to explore RSVIS.

\section{Methodology}
In this section, we commence by describing the construction of the two RSVIS datasets. Subsequently, we present a comprehensive overview of our method, illustrated in Fig.~\ref{fig:Overview}, and the specific network structure design for both the video-level and instrument-level embedding extraction branches, which can significantly complement the appearance information of instruments, thereby enhancing the segmentation accuracy. Next, we introduce the devised GRM, which exploits graph learning techniques to capture the inter-relationships among separate modalities. This enables the network to enhance the embedding of instrumental features referred to in the prompt text. The training and implementation details are presented at the end.

\begin{figure*}[h]
    \centering
    \includegraphics[width=1\textwidth]{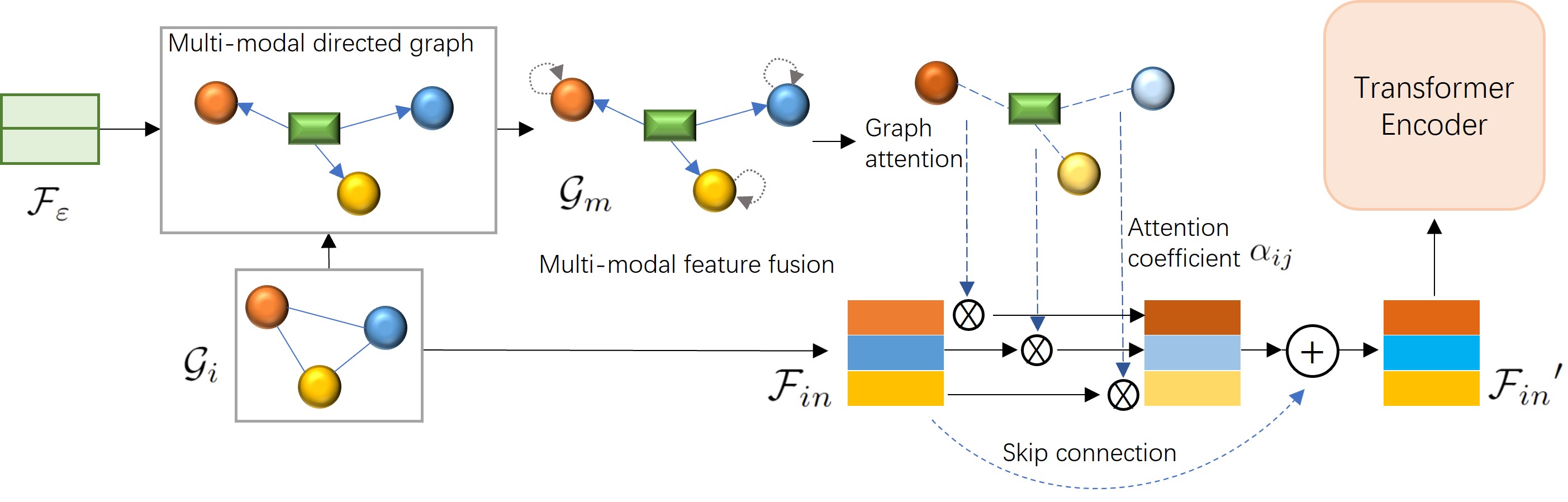}
    \vspace{-7mm}
    \caption{Detailed structure of the proposed Graph-based Relation-aware module. The circular nodes in the graph represent instrument-level visual embeddings, and the square node represents linguistic embedding.}
    \label{fig:GRM}
    \vspace{-3mm}
\end{figure*}

\begin{table}[t]
\centering
\caption{Detailed information regarding the proposed datasets. EndoVis-RS17 and EndoVis-RS18 are developed based on 2017 and 2018 Robotic EndoVis Dataset respectively.}
\label{data}
\resizebox{0.5\textwidth}{!}{%
\begin{tabular}{c|c|c|c}
\hline
\multirow{2}{*}{Dataset} &\multicolumn{3}{c}{Paired prompt textual description} \\
\cline{2-4} & instrument-text pairs & Word minimum & Word maximum \\
\hline
{\centering EndoVis-RS17} & {5133 pairs} & {7 words} & {9 words} \\
\hline
{\centering EndoVis-RS18} & {4711 pairs} & {3 words} & {6 words}  \\
\hline
\end{tabular}}
\begin{flushleft}
\textbf{Examples:}
"Suction is idle", "Bipolar forceps on the left is manipulating tool", "Prograsp forceps is grasping tissue on the right bottom"
\end{flushleft}
\vspace{-5mm}
\end{table}

\subsection{Dataset Construction}
We build the two RSVIS datasets based on the 2017 Robotic Instrument \cite{allan20192017} and 2018 Robotic Instrument \cite{allan20202018} Segmentation EndoVis Challenge. Information regarding the paired prompt textual description, is available in Table \ref{data}. 

We re-annotate the segmentation mask to ensure that multiple instruments of the same type can be distinguished in the identical surgical scenario. Subsequently, in collaboration with experienced surgeons, we scrupulously develop textual descriptions pertaining to the distinct instruments used during surgical procedures. These descriptions primarily encompass the activities performed by the instruments, and in certain scenarios, also include their spatial positioning within the surgical operation. This process is deemed necessary to effectively segment the instruments within complex surgical scenes and to differentiate between multiple instruments of the same type present within a single surgical setting. The active involvement of surgeons in determining the crucial aspects to be included in the labeling process underscores the professional and meticulous nature of our methodology. 
As evident from Table \ref{data}, the two datasets comprise 5133 and 4711 unique instrument-text pairs, respectively, ensuring a diverse and rich collection of sentences. 
In accordance with the essential information outlined in the doctor's recommendation, we endeavor to employ a wide range of prompt text formats, ensuring utmost flexibility in sentence styles, as exemplified in Table \ref{data}. 
We avoid evaluating the stapler due to the limited number of examples (only appearing in one video), following \cite{gonzalez2020isinet}. We statistically represent the distributions of the main words in both datasets, as shown in Fig.~\ref{fig:data}, reflecting the diversity of our sentences.

\subsection{Video-Instrument Synergistic Learning}
Fig. \ref{fig:Overview} shows the schematic illustration of our VIS-Net for RSVIS. 
The input of our VIS-Net includes a video clip $I=\{I_{t'},...,I_t\}$ with $T = t - t'$ frames, and a referring expression $\varepsilon=\{e_i\}_{i=1}^N$ with $N$ words, where $e_i$ is the $i$-th word in the text.
For the output, our VIS-Net predicts $T$ frame binary segmentation masks of the referred object $S=\{S_{t'},...,S_t\}$. First, feature extraction is performed on the input clips with both video level and instrument level, as well as text expression. Then, we devise a GRM to model the associations between the texts and the instruments, and the updated features are then concatenated with the text features and video-level embedding to form a multi-modal embedding $\mathcal{F}_m=\{\mathcal{F}_v,{\mathcal{F}_{in}}',\mathcal{F}_\varepsilon\}$ and fed to the Transformer encoder. After that, the corresponding masks and reference prediction sequences are generated via the Transformer encoder, the Transformer decoder, and FPN for predicting the final segmentation result. 

\textbf{Video-Instrument Visual and Linguistic Embedding Extraction.} 
For visual embeddings, VIS-Net complements more detailed information on the appearance of the instrument by performing the two-level embedding extractions. The video-level embeddings are extracted from the video clip $I$ by the spatio-temporal encoder. A suitable spatio-temporal encoder for the task requires extracting both motion and appearance information in the video. Previous referring segmentation works \cite{gavrilyuk2018actor,liu2021cross} using pre-trained I3D \cite{carreira2017quo} for instance segmentation (fine-detail task) is suboptimal due to spatial misalignment caused by temporal downsampling, as I3D is intended for classification. We utilize a recent advance, i.e., Video Swin Transformer \cite{liu2022video}, which is designed for dense predictions in mind, to process the video clip $I=\{I_{t'},...,I_t\}$, obtaining $\mathcal{F}_v$. Its shifted window-based self-attention mechanism can extract fine-grained and semantically rich features on different image patches and capture contextual information effectively.
Meanwhile, for instrument-level information extraction, video frames are first input into the object detection network, where we empirically utilize Yolov5 \cite{yolo} to detect instruments. Given the obvious gap between the instrument appearance and the surgical background, the detection network can effectively detect almost all instruments in the surgical scene. Then, we perform feature extraction for the detected areas through the deep model (i.e., ResNet \cite{he2016deep}) to obtain a more fine-grained appearance knowledge of different instruments $\mathcal{F}_{in}$. Simultaneously, the linguistic embeddings $\mathcal{F}_{\varepsilon}$ are extracted from the referring expression $\varepsilon$ by using a Transformer-based text encoder \cite{liu2019roberta}. Ultimately, the video-level visual embedding and linguistic embeddings are linearly projected to a shared embedding space with the same dimension. We further augment the instrument-level visual embedding by constructing a scene graph $\mathcal{G}_i$ and designing GRM to encourage to attain the text-guided instrument knowledge.

\subsection{Graph-based Relation-aware Module}
Given the multi-modal nature of the RSVIS task, it is pivotal to focus on the instrument's appearance mentioned in the text and ensure precise boundary segmentation during implementation. Therefore, establishing a pinpoint relationship between the text and image modalities is crucial for improving the visual representation of the referred surgical instrument. GRM is proposed to do this, inspired by the powerful capability of graph neural networks to model non-Euclidean relationships among entities \cite{long2021relational}. 
Concretely, we first construct a multi-modal directed graph, denoted as $\mathcal{G}_m=\{\mathcal{V}_m,\mathcal{E}_m,\mathcal{R}_m\}$ with nodes $v_i \in \mathcal{V}_m$ and edges $(v_j, r, v_i) \in \mathcal{E}_m$, where $r \in \mathcal{R}$ is a relation. Within our GRM, the relation $r \in \mathcal{R}$ denotes the directed one-way linkage, illustrating the flow of information from the text node to the visual nodes $(v_j \rightarrow v_i)$, thereby establishing the reference relationship.
The node entities $v_j$ and $\mathcal{V}_m$ correspond to text description and instrument-level visual information, whose associated features consist of linguistic embedding $\mathcal{F}_{\varepsilon}$ and the visual embeddings $\mathcal{F}_{in}$ from the instrument graph $\mathcal{G}_i$. Notably, the $\mathcal{F}_{\varepsilon}$ in Fig. \ref{fig:GRM} represents a multi-dimensional language embedding, and this multi-dimensional embedding is employed as a text node $v_j$ to construct the multi-modal graph $\mathcal{G}_m$.

As text and image information are essentially two types of modalities, there are domain gaps in the embedding distribution. We propose to first update the multi-modal graph by letting the two types of embeddings interact with each other. Since there are multiple instrument visual nodes and only one node is associated with the language node, the language node is vulnerable to contamination by extraneous visual nodes that are not referred by the current language node if the feature updating of connected nodes is performed by an undirected graph. Therefore, we propose to leverage the directed graph learning scheme to update the node embeddings. In our directed graph, only the text message propagates to the instrument's visual nodes, while its own embedding remains unchanged:
\begin{equation}
    F(\mathcal{V}_m,\widetilde{A})= ReLU (\widetilde{A} \mathcal{V}_m W_\alpha),
\end{equation}
where $\widetilde{A}$ represents the adjacency matrix containing the directed edge information with self-connection operation added. $\mathcal{V}_m$  includes the two-type node embeddings, and $W_\alpha$ is a layer-specific trainable weight matrix. 
Subsequent ablation experiments will also prove the effectiveness of this design. 
After minimizing the modality gap, we propose to augment visual nodes under the guidance of the relation matrix calculated from $\mathcal{G}_m$:
\begin{equation}
     \alpha_{ij} = softmax_j(e_{ij}) = \frac{exp(\Phi(W_\beta v_i,W_\beta v_j))}{\sum_{k \in \mathcal \mathcal{N}_i}exp(\Phi(W_\beta v_k,W_\beta v_j))},
\end{equation}
where $e_{ij}$ represents the edge $(v_i, r, v_j)$ of graph $\mathcal{G}_m$. Here, $v_j$ specifically refers to the text node while $v_i$ represents the visual node, and ${N}_i$ includes all visual nodes connected with $v_j$. 
$W_\beta$ is a shared trainable weight matrix. $\Phi$ converts two vectors into a scalar as the attention coefficient (single-layer feedforward neural network). 
Intuitively, the relation weight $\alpha_{ij}$ for node $v_i$ is obtained by normalizing its similarity over all the possibilities connected to $v_j$ via an attentional behavior by Softmax computation. Then, we employ the calculated coefficient $\alpha_{ij}$ to add the attended cues to augment the visual embedding $f_{in}\in\mathcal F_{in}$ in the instrument graph $\mathcal{G}_i$. We keep the original information by resuming it through a shortcut connection:
\begin{equation}
   {F_{in}}' = {F_{in}}+\alpha_{ij} \circ F_{in},
\end{equation}
where $\circ$ represents the Hadamard product.

\begin{table*}[ht]
\centering
\caption{Quantitative comparison of our network and state-of-the-art methods on the EndoVis-RS17 Dataset.}
\label{tab1}
\resizebox{0.8\textwidth}{!}{%
\begin{tabular}{c|c c c c c|c c|c}
\hline
\multirow{2}{*}{Method} & \multicolumn{5}{c|}{Precision} & \multicolumn{2}{c|}{IoU} & \multirow{2}{*}{mAP}\\
 & P@0.5 & P@0.6 & P@0.7 & P@0.8 & P@0.9 & Overall & Mean & \\
\hline
URVOS \cite{seo2020urvos} & 65.9 & 61.2 & 56.5 & 48.8 & 25.8 & 57.7 & 60.8 & 47.9 \\
CMPC-V \cite{liu2021cross} & 66.8 & 62.1 & 58.2 & 48.8 & 26.1 & 59.4 & 61.8 & 48.8 \\
LBDT \cite{ding2022language} & 71.0 & 64.5 & 60.1 & 52.5 & 29.6 & 61.0 & 65.6 & 51.9 \\
MTTR \cite{botach2022end} & 71.7 & 64.8 & 60.9 & 52.9 & 30.2 & 61.3 & 66.3 & 52.2 \\
\hline
Our VIS-Net & \textbf{76.1} & \textbf{66.4} & \textbf{62.3} & \textbf{54.6} & \textbf{32.2} & \textbf{65.5} & \textbf{70.1} & \textbf{53.8} \\
Gain & $\uparrow$4.4 & $\uparrow$1.6 & $\uparrow$1.4 & $\uparrow$1.7 & $\uparrow$2.0 & $\uparrow$4.2 & $\uparrow$3.8 & $\uparrow$1.6 \\
\hline
\end{tabular}}
\vspace{-3mm}
\end{table*}

\begin{table*}[h]
\centering
\caption{Quantitative comparison of our network and state-of-the-art methods on the EndoVis-RS18 Dataset.}
\label{tab2}
\resizebox{0.8\textwidth}{!}{%
\begin{tabular}{c|c c c c c|c c|c}
\hline
\multirow{2}{*}{Method} & \multicolumn{5}{c|}{Precision} & \multicolumn{2}{c|}{IoU} & \multirow{2}{*}{mAP}\\
 & P@0.5 & P@0.6 & P@0.7 & P@0.8 & P@0.9 & Overall & Mean & \\
\hline
URVOS \cite{seo2020urvos} & 69.3 & 65.0 & 59.3 & 51.3 & 31.5 & 66.7 & 63.4 & 52.0 \\
CMPC-V \cite{liu2021cross} & 70.3 & 66.0 & 62.4 & 55.1 & 34.7 & 67.3 & 66.1 & 54.3 \\
LBDT \cite{ding2022language} &  76.5  & 72.0 & 65.7 & 56.7 & 34.0 & 71.9 & 70.2 & 57.5 \\
MTTR \cite{botach2022end} & 77.1 & 72.2 & 66.0 & 57.6 & 34.8 & 72.2 & 70.8 & 58.0 \\
\hline
Our VIS-Net & \textbf{80.2} & \textbf{75.2} & \textbf{68.5} & \textbf{60.2} & \textbf{36.1} & \textbf{74.2} & \textbf{72.3} & \textbf{60.1} \\
Gain & $\uparrow$3.1 & $\uparrow$3.0 & $\uparrow$2.5 & $\uparrow$2.6 & $\uparrow$1.3 & $\uparrow$2.0 & $\uparrow$1.5 & $\uparrow$2.1 \\
\hline
\end{tabular}}
\vspace{-3mm}
\end{table*}

\subsection{Model Training and Implementation Details}
In the next step, the two-level visual embeddings $\{\mathcal{F}_{v},{\mathcal{F}_{in}}'\}$ are flattened and concatenated with the text information $\mathcal{F}_{\varepsilon}$, and then formed as a set of multi-modal sequences. 
We feed the concatenation result into the Transformer’s encoder layers \cite{vaswani2017attention}, where the textual embedding and the visual embeddings can exchange information. 
Then, Transformer decoder layers receive $N_q$ instrument queries (represented by the unique color and shape in Fig.\ref{fig:Overview}) per input frame to retrieve entity-related information from the multi-modal sequences. The queries corresponding to different frames share identical trainable weights and are trained to focus on the same instance within the video. These queries, referred to as queries belonging to the same instance sequence, enable seamless tracking of individual object instances throughout the video. 
For each output instance sequence, it generates a corresponding mask sequence using an FPN-like decoder~\cite{lin2017feature}. 
We then select an instance sequence that has a strong association with the input text description, and then utilize its segmentation result as the final segmentation result of our VIS-Net.
Moreover, we follow MTTR \cite{botach2022end} to compute the loss function of our segmentation result, mainly inlcudes the pair-wise matching loss computed by using Hungarian algorithm \cite{kuhn1955hungarian} and the Dice coefficients loss for accurate segmentation.

All experiments are conducted on PyTorch and trained on an NVIDIA GeForce RTX 3090 GPU with a 24 GB memory. For the spatio-temporal encoder, we use the pretrained smallest (“tiny”) Video Swin Transformer \cite{liu2022video}, and we use a window size of 3 consecutive annotated frames during training. 
Each frame is resized so that the shorter side is at least 360 pixels and the longer side is at most 640 pixels. For the text encoder, we employ RoBERTa~\cite{liu2019roberta}. For object detection, BCEWithLogits loss and CIoU loss are applied to compute the loss function. 
As for detecting instrument areas, all image patches are resized to 224×224 before we pass them into ResNet18. 
The decoder layers are fed with a set of $N_q = 50$ object queries per frame. 
All models are trained with 100 epochs and AdamW optimizer~\cite{loshchilov2017decoupled}  and a batch size of 5.
The initial learning rate of the main model is 0.0001, and then reduced by a factor of 0.6 at the $20th$ and $40th$ epochs.

\begin{figure*}[h!]
    \centering
    \includegraphics[width=0.93\textwidth]{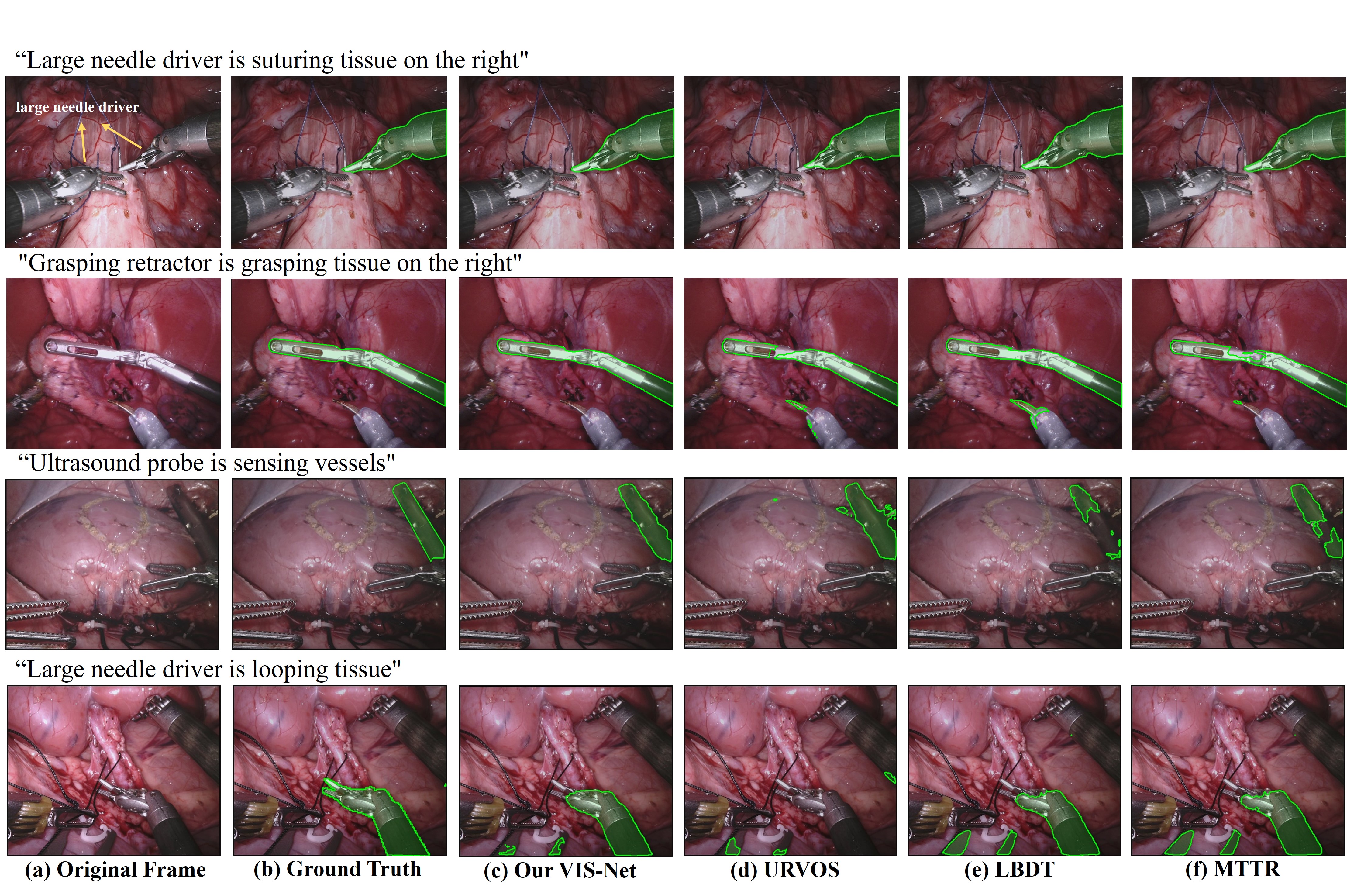}
    \vspace{-3mm}
    \caption{Visually comparisons of referring instrument segmentation results produced by our VIS-Net and three SOTA methods. Apparently, our VIS-Net can more accurately segment the instrument regions referred by the corresponding words. The top two columns are from EndoVis-RS17 and the bottom two columns are from EndoVis-RS18.}
    \label{fig:demo}
    \vspace{-3mm}
\end{figure*}

\section{Experiments and Results}

\subsection{Experimental Setting and Evaluation Metrics}
Extensive experiments are performed on two tailored RSVIS datasets to rigorously evaluate the effectiveness of the proposed VIS-Net. We divide the original surgical videos of the EndoVis-RS17 dataset into two sets: a validation set containing 8th-13th video sequences, and a training set with the remaining 9 video sequences (exclude the corresponding video in training data when evaluating different validation videos). For the EndoVis-RS18 dataset, video sequences 2nd, 5th, 9th and 15th are employed as the validation set, as well as the other 10 video sequences for network training. This data division strategy is implemented to ensure a balanced representation of all instruments in both the training and validation sets.

Following previous works~\cite{botach2022end,ding2022language}, we adopt a set of widely employed metrics to facilitate quantitative evaluation. These metrics include Precision@K, Overall IoU, Mean IoU, and mAP (mean Average Precision) over the IoU (Intersection over Union) range of 0.50 to 0.95 with an interval of 0.05. IoU measures the overlap between the predicted region and the ground truth by calculating the ratio of the intersection area to the union area of the two regions. Precision@K assesses the proportion of test samples with IoU scores exceeding the threshold K. The mAP calculates the average precision across different IoU thresholds, ranging from 0.50 to 0.95 with an interval of 0.05. Generally, a superior method is expected to exhibit higher scores across all these quantitative metrics.

\begin{figure}[t]
    \centering
    \includegraphics[width=0.48\textwidth]{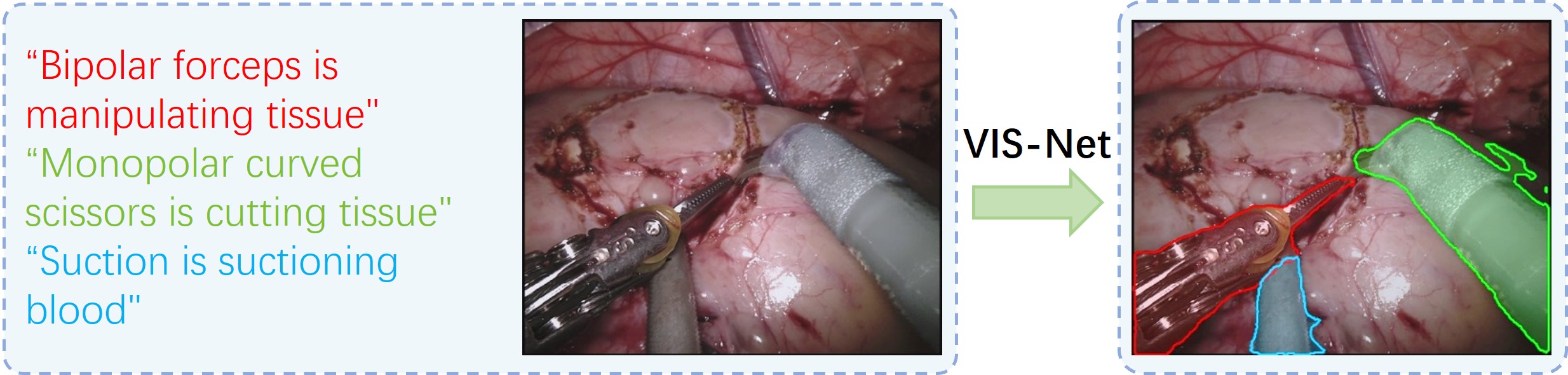}
    \vspace{-1mm}
    \caption{More segmentation results of our method. Our method can effectively segment the corresponding surgical instruments referred by three different sentences.}
    \label{fig:demo2}
    \vspace{-3mm}
\end{figure}

\subsection{Comparison with State-of-the-arts}
We conduct a comparison between our method and state-of-the-art (SOTA) methods to evaluate their performance, including URVOS \cite{seo2020urvos}, CMPC-V \cite{liu2021cross} LBDT \cite{ding2022language}, and MTTR \cite{botach2022end}.
Table~\ref{tab1} and Table~\ref{tab2} report the quantitative results of our network and four compared methods on the EndoVis-RS17 and EndoVis-RS18 datasets. 
From the analysis of Table~\ref{tab1} and Table~\ref{tab2}, it becomes evident that MTTR exhibits superior performance in terms of the evaluated metrics when compared to the other three methods under consideration. Furthermore, it is worth noting that our model consistently surpasses MTTR by a substantial margin across all quantitative metrics, emphasizing its remarkable performance superiority. 
Our results of the P@0.5, Overall IoU, and mAP are 76.1\%, 65.5\%, and 53.8\% for the EndoVis-RS17 datasets and 80.2\%, 74.2\%, and 60.1\% for the EndoVis-RS18 datasets, respectively. This observation highlights the exceptional ability of our network to achieve a superior level of accuracy in referring instrument segmentation within surgical videos. The last row of Table \ref{tab1} and Table \ref{tab2} also demonstrates the numerical improvement in different metrics compared to the SOTA methods.

Fig.\ref{fig:demo} visually compares the referred instruments segmentation results of some input surgical video frames with the corresponding textual descriptions.
We can observe that VIS-Net demonstrates a higher level of precision in instrument segmentation, yielding results that align most closely with the ground truth (GT) across a wide range of surgical video frames, encompassing both complex and simple scenes. However, other compared methods tend to produce incomplete instrument segmentation results (see LBDT and MTTR at the 3rd row, and URVOS at the 1st and 4th row) or tend to include other instruments, which are not related to the input sentence; see the segmentation results of URVOS, LBDT and MTTR at the 2nd and 4th rows. 
Moreover, Fig.\ref{fig:demo2} demonstrates our segmentation results with multiple input sentences. Apparently, our approach can effectively segment the corresponding surgical instruments, which are referred by the input sentences.

\subsection{Analysis of Key Components}
To validate the effectiveness of various key components in our proposed method, we conduct ablation experiments, resulting in four configurations: (i) Baseline: we train the pure MTTR backbone network as the baseline model; (ii) M1: this involves training the baseline model along with the instrument-level branch; (iii) M2: building upon the M1 configuration, we incorporated Graph attention to enhance the instrument-level embedding; (iv) Ours VIS-Net: this configuration encompasses both the video-level and instrument-level branches, along with the complete GRM (Graph attention and $G_m$).
\begin{table*}[ht]
\caption{Ablation study of the proposed methods on the EndoVis-RS17 Dataset.}\label{tab3}
\resizebox{\textwidth}{!}{%
\begin{tabular}{c|c|c c|c c c c c|c c|c}
\hline
\multirow{2}{*}{Method} & \multirow{2}{*}{Instrument branch} & \multicolumn{2}{c|}{GRM} & \multicolumn{5}{c|}{Precision} & \multicolumn{2}{c|}{IoU} & \multirow{2}{*}{mAP}\\
 & & Graph attention & $G_m$ & P@0.5 & P@0.6 & P@0.7 & P@0.8 & P@0.9 & Overall & Mean & \\
\hline
Baseline &  &  &  & 71.7 & 64.8 & 60.9 & 52.9 & 30.2 & 61.3 & 66.3 & 52.2 \\
M1 & \usym{2713} &  &  & 73.4 & 65.0 & 61.6 & 53.3 & 30.4 & 63.9 & 68.2 & 53.0 \\
M2 & \usym{2713}  &\usym{2713} & & 74.8 & 66.0 & 62.1 & 54.1 & 30.8 & 64.4 & 69.1 & 53.1 \\
Ours VIS-Net & \usym{2713} & \usym{2713} &\usym{2713} & \textbf{76.1} & \textbf{66.4} & \textbf{62.3} & \textbf{54.6} & \textbf{32.2} & \textbf{65.5} & \textbf{70.1} & \textbf{53.8} \\
\hline
\end{tabular}}
\vspace{-3.3mm}
\end{table*}

\begin{table*}[t]
\caption{Ablation study of the proposed methods on the EndoVis-RS18 Dataset.}\label{tab4}
\resizebox{\textwidth}{!}{%
\begin{tabular}{c|c|c c|c c c c c|c c|c}
\hline
\multirow{2}{*}{Method} & \multirow{2}{*}{Instrument branch} & \multicolumn{2}{c|}{GRM} & \multicolumn{5}{c|}{Precision} & \multicolumn{2}{c|}{IoU} & \multirow{2}{*}{mAP}\\
 & & Graph attention & $G_m$ & P@0.5 & P@0.6 & P@0.7 & P@0.8 & P@0.9 & Overall & Mean & \\
\hline
Baseline &  &  &  & 77.1 & 72.2 & 66.0 & 57.6 & 34.8 & 72.2 & 70.8 & 58.0 \\
M1 & \usym{2713} &  &  & 77.8 & 72.7 & 67.2 & 57.9 & 35.9 & 72.4 & 70.6 & 58.8 \\
M2 & \usym{2713}  &\usym{2713} & & 78.4 & 73.9 & 68.0 & 59.7 & 35.9 & 73.3 & 70.9 & 59.7 \\
Ours VIS-Net & \usym{2713} & \usym{2713} &\usym{2713} & \textbf{80.2} & \textbf{75.2} & \textbf{68.5} & \textbf{60.2} & \textbf{36.1} & \textbf{74.2} & \textbf{72.3} & \textbf{60.1} \\
\hline
\end{tabular}}
\vspace{-1mm}
\end{table*}

\begin{figure*}[h]
    \centering
    \includegraphics[width=1\textwidth]{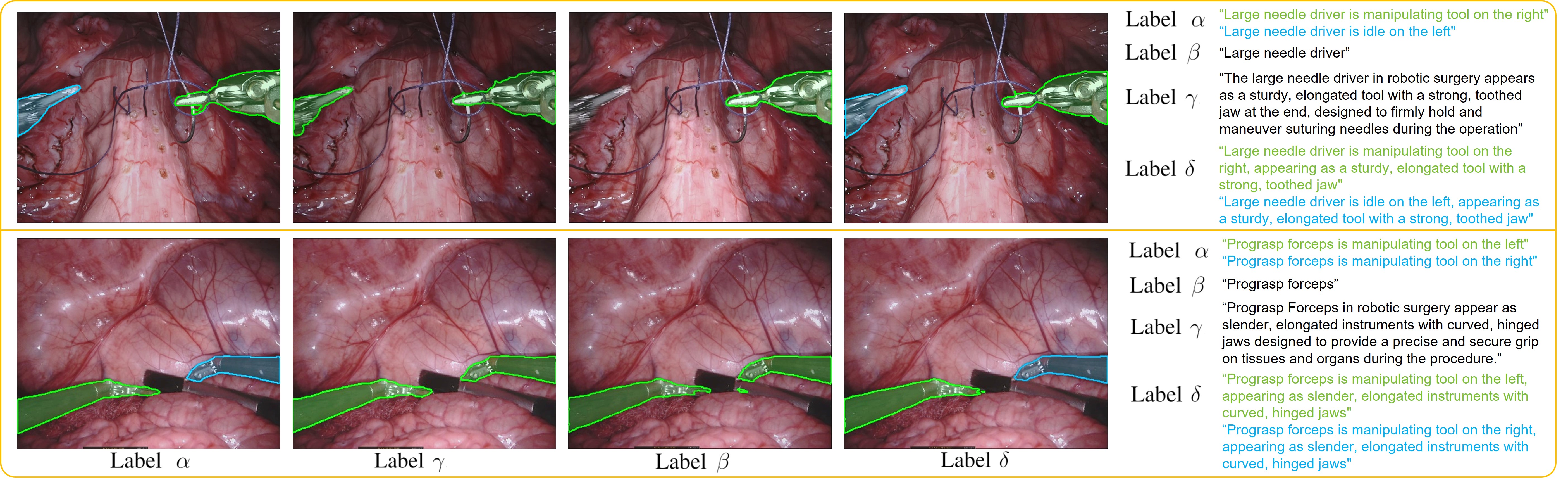}
    \caption{Comparison of visual segmentation results based on different text descriptions. Label $\alpha$ and Label $\delta$ accurately distinguish multiple instruments with the same type, while the other two fail. The Label $\beta$ misses the instrument on the left shown in the first row, and Label $\gamma$ cannot distinguish between the two instruments.}
    \label{fig:demo4}
    \vspace{-4mm}
\end{figure*}

Table~\ref{tab3} and Table~\ref{tab4} present the results of the ablation experiments conducted on the two datasets. Remarkably consistent outcomes are observed across both sets of experiments. The inclusion of instrument-level embeddings in M1 demonstrated improved performance compared to the "Baseline" model, as evidenced by various evaluation metrics. Subsequently, M2 outperformed M1, indicating that the incorporation of graph attention (Eq. (2)) within our GRM significantly enhanced the segmentation performance. 
Importantly, the notable improvements in metric results achieved by our VIS-Net compared to M2 serve as compelling evidence for the efficacy of incorporating $G_m$ (Eq. (1)) for calculating attention coefficients within our GRM. This operation plays a pivotal role in achieving more accurate segmentation of instruments referred to in the surgcial video. It serves as a testament to the effectiveness of the multi-modal graph $G_m$ as a crucial connecting link bridging the two modalities. 

\begin{table*}[h]
\centering
\caption{Experiments on the EndoVis-RS17 Dataset based on multiple formats of description texts.}\label{tab5}
\resizebox{0.9\textwidth}{!}{%
\begin{tabular}{c|c c c c|c c c c c|c c|c}
\hline
\multirow{2}{*}{Method}  & \multicolumn{4}{c|}{Label} & \multicolumn{5}{c|}{Precision} & \multicolumn{2}{c|}{IoU} & \multirow{2}{*}{mAP}\\
 & $\beta$ & $\gamma$ & $\alpha$ & $\delta$ & P@0.5 & P@0.6 & P@0.7 & P@0.8 & P@0.9 & Overall & Mean & \\
\hline
LBDT \cite{ding2022language} &\usym{2713}  &  &  &  & 47.6 & 40.4 & 37.7 & 33.3 & 21.9 & 39.2 & 47.6 & 33.6 \\
LBDT \cite{ding2022language} &  & \usym{2713}  &  &  & 49.1 & 42.4 & 40.1 & 35.5 & 22.4 & 39.7 & 48.6 & 35.3 \\
LBDT \cite{ding2022language} &  & & \usym{2713}   &  & 71.0 & 64.5 & 60.1 & 52.5 & 29.6 & 61.0 & 65.6 & 51.9 \\
LBDT \cite{ding2022language} &  & &  & \usym{2713} &  74.1 & 63.8 & 59.8 & 50.6 & 29.6 & 64.4 & 68.1 & 35.6  \\
MTTR \cite{botach2022end} &\usym{2713}  &  &  &   & 49.1 & 42.8 & 40.1 & 35.1 & 23.8 & 41.1 & 49.4 & 35.7 \\
MTTR \cite{botach2022end} &  & \usym{2713}  &  &   & 49.6 & 42.8 & 40.5 & 36.0 & 22.8 & 40.0 & 48.9 & 35.7 \\
MTTR \cite{botach2022end} &  & & \usym{2713}  &   & 71.7 & 64.8 & 60.9 & 52.9 & 30.2 & 61.3 & 66.3 & 52.2 \\
MTTR \cite{botach2022end} &  & &  & \usym{2713} & 74.7 & 64.9 & 60.1 & 52.7 & 31.5 & 64.8 & 69.1 & 52.3 \\
\hline
Our VIS-Net &  \usym{2713} & &  &  & 50.4  & 44.3 & 41.8 & 37.5 & 23.0 & 40.7 & 49.5 & 36.8 \\
Our VIS-Net &   &\usym{2713} & & & 51.6  & 46.3 & 43.5 & 39.7 & 26.7 & 43.2 & 50.4 & 39.6 \\
Our VIS-Net &  &  & \usym{2713}  & & 76.1 & 66.4 & 62.3 & 54.6 & 32.2 & 65.5 & 70.1 & 53.8 \\
Our VIS-Net &  &  &  & \usym{2713} &\textbf{76.6} & \textbf{67.4} & \textbf{63.3} & \textbf{55.0} & \textbf{33.3} & \textbf{66.5} & \textbf{70.2} & \textbf{54.4} \\
\hline
\end{tabular}}
\vspace{-3mm}
\end{table*}

\subsection{Analysis of Text Formatting}
In this section, we will further explore the significance and advantages of RSVIS, while experimenting with different formats of text on the EndoVis-RS17 dataset to verify the performance of our methodology. It should be noted that the text annotated with the help of our surgeons is named Label $\alpha$ in this session, text containing only the names of the instruments is called Label $\beta$ (such as bipolar forceps and prograsp forceps), and text descriptions generated using GPT-4 \cite{sanderson2023gpt} is called Label $\gamma$. The concrete method of generating corresponding text for GPT-4 is to enter a template as, "Please describe the appearance of [specific instrument] in robotic surgery, and change the description to a phrase with subject". By employing this approach, it produces text like "Prograsp Forceps in robotic surgery appear as slender, elongated instruments with curved, hinged jaws designed to provide a precise and secure grip on tissues and organs during the procedure." as shown in Fig \ref{fig:demo4}. 
Taking this a step further, we endeavor to merge Label $\alpha$ and Label $\gamma$ to craft a more sophisticated Label $\delta$. 
We selecte two representative methods (LBDT \cite{ding2022language} and MTTR \cite{botach2022end}) and our VIS-Net to conduct experiments based on text descriptions in these three formats as shown in Table \ref{tab5}.

One point to clarify is that our Label $\alpha$ and Label $\delta$ are fully RSVIS-compliant in both datasets and can segment any particular instrument, whereas Label $\beta$ and Label $\gamma$ are only RSVIS-compliant in some cases where there are not multiple instruments with the same type in one scenario. Because Label $\beta$ and $\gamma$ cannot segment a particular instrument when there are multiple instruments with the same type as in Fig.~\ref{fig:demo4}. It is essentially more like a traditional instrument segmentation using text to enhance. Combined with Table \ref{tab5} it can also be seen that they both show a similarly large decline in performance. Our method also outperforms other methods, probably owing to the instrument-level embedding. Combining with Table~\ref{tab5} reveals that Label $\delta$ indeed leads to a performance improvement. Our method still achieves the best results, which demonstrates the robustness of our method and the flexibility of accepting prompt sentences.

\section{Discussion}
Surgical instrument segmentation is a fundamental challenge in the field of Computer-Assisted Interventions (CAI), with wide-ranging implications and applications \cite{sestini2023fun,garcia2021image,colleoni2022ssis}. Accurately identifying and delineating the boundaries of surgical instruments within a surgical scene holds significant potential for various clinical applications \cite{wang2022rethinking,shen2023branch}. In fact, it already serves as a crucial component or is poised to become a vital element in numerous essential clinical applications within the domain of surgical data science. While significant advancements have been made in surgical instrument segmentation, these approaches encounter limitations when confronted with scenarios involving multiple instruments of the same type within a surgical scene. Therefore, this paper introduces a novel problem in the realm of video-language learning, referred to as Referring Surgical Video Instrument Segmentation (RSVIS). The objective of RSVIS is to automatically identify and segment surgical instruments based on the provided language expression. By addressing this problem, we aim to achieve more fine-grained instrument segmentation while enabling natural human-computer interaction. This research has the potential to make significant contributions to the advancement and integration of context-aware intelligent systems in the operating room and surgical education domain.

Nevertheless, unlike the extensively studied field of natural video referring segmentation\cite{liu2021cross,ding2022language}, there has been a lack of research specifically focusing on the RSVIS task. This is primarily attributed to the inherent complexity of surgical videos, where instruments may exhibit strikingly similar appearances, making accurate segmentation a challenging endeavor. Considering the distinctive attributes of surgical videos, we develop the Video-Instrument Synergistic Network, which leverages instrument-level appearance information to enhance boundary segmentation accuracy. Additionally, drawing inspiration from the GNN's capability to capture non-Euclidean distances, we employ graph learning techniques to model the multi-modal relationships. This approach facilitates the enhancement of instrument embedding referred to in the text to facilitate the final segmentation accuracy.

Another significant challenge in this research domain is the lack of availability of appropriately annotated video datasets. The majority of existing datasets are primarily designed for conventional instrument type segmentations or semantic segmentation tasks, which do not align with the specific requirements of our newly proposed RSVIS task. To overcome this challenge, we take the initiative to create two well-annotated RSVIS datasets with the guidance of surgeons using surgical videos from the widely recognized EndoVis Challenge 2017 and 2018 datasets. These newly constructed datasets are specifically tailored to fulfill the analytical and practical requirements of RSVIS research. Importantly, we are committed to making these datasets publicly available, thereby facilitating further advancements and encouraging researchers to contribute to the field of RSVIS.

Our proposed method has undergone rigorous validation on the aforementioned datasets, demonstrating its superiority over the state-of-the-art approaches. The ablation studies demonstrate the effectiveness of our designed method. The limitation of our work is that we have currently annotated the data with the help of surgeons, so the richness of the sentences may be limited. To alleviate this limitation, we have explored the application of GPT-4 \cite{sanderson2023gpt} to generate more diverse and comprehensive sentence descriptions. Additionally, we conduct experiments to compare the impact of various text annotations (including Label $\alpha$, Label $\beta$, Label $\gamma$ and Label $\delta$) on performance. As evident in Table \ref{tab5} and Fig.~\ref{fig:demo4}, Our approach achieves the best performance across a variety of text formats. These endeavors are conducted to substantiate the importance of RSVIS and to showcase the resilience of our approach. Moreover, these efforts aim to highlight the robustness of our approach and the flexibility of the sentences that could be accepted.

\section{Conclusion}
In this paper, we are the first to collect and annotate two high-quality datasets from RSVIS. Based on these surgical video datasets, we further devise a Video-Instrument Synergistic network to capture both video-level and instrument-level information to improve the segmentation accuracy. 
Furthermore, we devise a Graph-based Relation-aware Module to learn the relationship between the input surgical video and the input texts. Experimental results demonstrate that our network clearly outperforms state-of-the-art methods. Meanwhile, Detailed ablation experiments and exploratory experiments and discussions on the format of the text have also been carried out and have demonstrated the significance of RSVIS and the robustness of our method. In the future, we plan to consider incorporating other modality information, such as kinematics to facilitate RSVIS.

\bibliographystyle{IEEEtran}
\bibliography{ref}

\end{document}